\documentclass[11pt]{article}
\usepackage{amsmath}

\usepackage[final]{acl}

\usepackage{times}
\usepackage{latexsym}
\usepackage{todonotes}
\usepackage{multirow}

\usepackage[T1]{fontenc}

\usepackage[utf8]{inputenc}

\usepackage{microtype}

\usepackage{inconsolata}

\usepackage{graphicx}

\usepackage{listings}
\usepackage{cuted}

\lstdefinestyle{prompt}{
    basicstyle=\ttfamily\footnotesize,
    breaklines=true,
    breakatwhitespace=false,
    columns=fullflexible,
    keepspaces=true,
    showstringspaces=false,
    frame=none,
    xleftmargin=0pt,
    xrightmargin=0pt,
    aboveskip=4pt,
    belowskip=6pt
}

\title{From Utterances to Networks: \\ Modelling Slang Adoption and Diffusion Across Subreddits}

\author{
 \textbf{Xiaoning Wang\textsuperscript{1}},
 \textbf{Ted Underwood\textsuperscript{1}},
 \textbf{Zhewei Sun\textsuperscript{2}}
\\[5pt]
\textsuperscript{1}School of Information Sciences, University of Illinois Urbana-Champaign
\\
\texttt{\fontfamily{pcr}\selectfont xw109@illinois.edu, tunder@illinois.edu}
\\[5pt]
\textsuperscript{2}Toyota Technological Institute at Chicago
\\
\texttt{\fontfamily{pcr}\selectfont zsun@ttic.edu}
\\
}

\begin{document}
\maketitle
\begingroup
\renewcommand{\thefootnote}{}
\footnotetext{Code and data are available at: \url{xiaoningwang.ca/projects/slang}.}
\endgroup

\begin{abstract}

\end{abstract}
Adoption and diffusion of neologisms in online communities have received renewed attention in recent years. As internet slang terms such as \textit{APT}, referring to a K-pop song, and phrases such as \textit{Canon Event} meaning an embarrassing but pivotal event, go viral online, it becomes increasingly important to understand the mechanisms that contribute to their success. Prior studies have often explained slang diffusion either from the perspective of social interaction or from the linguistic properties of the slang itself, but rarely from both perspectives together. One major obstacle has been the high cost of annotating slang usage in large-scale online communication. Recent advances in large language models (LLMs), however, make it possible to use them as scalable annotators for such tasks.
In this study, we first curate a human-annotated benchmark to evaluate LLM performance in detecting slang usage in real-world Reddit communication. We then leverage LLM-based annotations to model slang adoption and diffusion. Our results show that, at the slang-diffusion level, bridging positions in the network facilitate the spread of slang, whereas greater local connectedness has the opposite effect. At the user level, both network structure and semantic context influence how quickly a user adopts slang.

\section{Introduction}

\begin{figure}[t]
  \centering
  \includegraphics[width=\columnwidth]{figures/benchmark-example-sketch-4.pdf}
  \caption{An example of an annotated Reddit utterance from r/asoiaf. <glossary-sense> wraps slang terms used in their sense according to the community glossary, while <non-glossary-sense> wraps term in the words' other senses. For example, in this subreddit, "book!" is a slang used to refer to a character as they appear in the book series A Song of Ice and Fire (e.g. "Book!Sansa").}
  \label{fig:fig1}
\end{figure}

Lexical change in social networks has received growing attention in recent years, as social interaction plays a central role in both the emergence of new lexical forms and the maintenance of existing ones \cite{doi:https://doi.org/10.1002/9781118335598.ch19}. A lexical innovation typically begins when a speaker, also known as an inventor, coins a new term or assigns a new meaning to an existing form. Through social interaction, this innovation becomes available to potential adopters, who may then take it up in their own language usages. At this stage, lexical innovations that are better suited to speakers’ communicative contexts are more likely to be taken up and reproduced. The diffusion of lexical innovations therefore depends not only on patterns of social interaction, but also on the characteristics of the innovations themselves. We therefore distinguish between network features, which capture how users encounter slang through interaction, and linguistic features, which capture properties of the slang usage contexts that may affect adoption.


Lexical innovation has also been conceptualized as a diffusion process analogous to social contagion. Prior work has modeled the spread of neologisms using epidemic models such as SIR \cite{Jiang2021-bk}, while more recent work has used network-based simulation models to characterize how lexical innovations propagate between speakers \cite{Ananthasubramaniam2024-gi}.

In most prior studies, they were primarily focused on network features \cite{Hamilton2017-cy, zhu-jurgens-2021-structure, Del_Tredici2018-vf} and very little on linguistic features beyond the word forms \cite{stewart-eisenstein-2018-making}. According to theory of innovation diffusion \cite{Rogers2003-mo}, semantics, in the case of lexical innovation, is one of the necessary conditions determining whether potential adopters use an innovation. Another key obstacle preventing existing studies from incorporating linguistic features is the difficulty of collecting utterances that contain slang, since slang terms commonly recycles existing word forms~\cite{warren92, eble12}. Existing studies relied heavily on dictionary-based detection \cite{zhu-jurgens-2021-structure, Del_Tredici2018-vf}, because identifying slang usages at scale requires either substantial human annotation or automated word sense disambiguation (WSD), which has historically fallen short of acceptable accuracy. Recent advances in large language models (LLMs) have made it feasible to use LLMs as judges to detect and disambiguate slang across large corpora of utterances \cite{sun-etal-2024-toward, wuraola-etal-2024-understanding}. In this work, we provide a benchmark for evaluating LLMs’ ability to detect slang in real-world Reddit conversations. We find that Gemma-4-E4B-it achieves up to 93.44\% precision in WSD, suggesting that it can serve as a reliable annotator.

By leveraging LLMs, we can incorporate both network and linguistic features into the analysis of informal lexical change on online social media. We treat slang terms on Reddit as lexical innovations and study how slang diffuses and how users adopt these terms within a community. We focus on 54 subreddits that maintained moderator-curated slang glossaries \cite{Lucy2021-kk}, a dataset that enables two complementary analyses. First, it allows us to study which network and linguistic features predict whether a slang term becomes widely adopted within a community, as evidenced by its inclusion in the glossary. Second, it allows us to identify slang terms coined in one subreddit and trace how they are subsequently adopted by members of that community. Our analysis shows that slang diffused by users with greater bridging capital is more likely to spread, while greater diversity in usage contexts may both facilitate diffusion and create hurdles for new users learning and adopting these terms.

In this paper, we make the following contributions: 1) A publicly available benchmark of 1,333 real Reddit utterances with human-annotated, subreddit-specific English slang usages across 23 subreddits, supporting the evaluation of LLM-based slang detection and WSD in naturalistic conversation. 2) An evaluation of how network and linguistic features contribute to slang dissemination and adoption with fixed effect model and survival analysis.

\section{Data}
\subsection{Reddit Corpus}

\begin{table*}[t]
  \centering
  \small
  \setlength{\tabcolsep}{7pt}
  \renewcommand{\arraystretch}{1.25}

  \begin{tabular}{llcccc}
    \textbf{Model} & \textbf{Task} &
    \textbf{Precision} & \textbf{Recall} & \textbf{F1} & \textbf{N} \\
    \hline
    \noalign{\vskip 3pt}

    \multirow{2}{*}{Llama-3.1-8B-Instruct}
      & Single-WSD
      & 85.11 & 88.03 & 86.55 & 1,853 \\
      & List-WSD
      & 62.07 & 67.27 & 63.19 & 1,333 \\[3pt]

    \hline
    \noalign{\vskip 3pt}

    \multirow{2}{*}{Ministral-3-8B-Instruct-2512}
      & Single-WSD
      & 93.05 & 87.83 & 90.36 & 1,853 \\
      & List-WSD
      & 83.84 & 81.79 & 82.10 & 1,333 \\[3pt]

    \hline
    \noalign{\vskip 3pt}

    \multirow{2}{*}{OLMo-3-7B-Instruct}
      & Single-WSD
      & 80.71 & 87.10 & 83.78 & 1,853 \\
      & List-WSD
      & 73.67 & 70.26 & 70.59 & 1,333 \\[3pt]

    \hline
    \noalign{\vskip 3pt}

    \multirow{2}{*}{Gemma-4-E4B-it}
      & Single-WSD
      & \textbf{93.44} & \textbf{93.44} & \textbf{93.44} & 1,853 \\
      & List-WSD
      & \textbf{84.59} & \textbf{84.71} & \textbf{83.98} & 1,333 \\[3pt]

    \hline
  \end{tabular}

  \caption{
    Benchmark results for slang detection on the Single-WSD and List-WSD tasks. For Single-WSD, the model is given a Reddit comment, a target term, and its corresponding glossary definition, and determines whether the term is used in its glossary sense. For List-WSD, the model is given a Reddit comment and the full slang glossary and identifies which glossary terms are used in the comment. The best result for each task and metric is shown in bold.    
  }
  \label{tab:benchmark-result}
\end{table*}

To analyze slang diffusion and adoption, we follow \citet{zhu-jurgens-2021-structure} in using comments posted to Reddit and treating each subreddit as a community. We use the ConvoKit package \cite{Chang2020-gr} as our source of the Reddit corpus, whose dataset spans ten years from 2008 to 2018, with the majority of data concentrated after 2013. 

We obtain the moderator-curated glossary from \citet{Lucy2021-kk}, who compiled slang lists from the corresponding moderator threads. Each entry contains the slang term, its subreddit, and a definition. Our analysis covers the 54 subreddits that are available on ConvoKit.

\subsection{Slang Detection Dataset}

To evaluate LLM performance on slang detection in real Reddit comments, we curated a benchmark of 1,333 human-annotated comments across 23 subreddits. Unlike prior datasets that are based on subtitles rather than naturally occurring subreddit conversations \cite{sun-etal-2024-toward}, focus primarily on binary slang/non-slang classification \cite{Aloraini2025-wd}, or are designed for related tasks such as slang comprehension \cite{mei-etal-2024-slang} and sentiment analysis \cite{wuraola-etal-2024-understanding}, our benchmark evaluates model performance on real utterances from subreddit conversations. In this setting, a single input may contain multiple slang candidates, each of which may be used either in its sense in glossary or not. Figure~\ref{fig:fig1} shows one annotated example. Each entry is accompanied by metadata, including the source subreddit, the corresponding subreddit glossary, and the surrounding conversational context of the comment. 

In practical use, we aim to provide real Reddit conversations directly to an LLM and ask it to identify slang usages without additional preprocessing. However, slang detection is challenging not only because slang often reuses existing word forms, requiring the model to distinguish slang senses from literal senses, but also because processing long conversational contexts can be difficult for LLM-based annotation. In contrast, newly coined slang forms are comparatively straightforward to detect. Therefore, we decompose slang detection into WSD and term retrieval to evaluate model performance more diagnostically. In \textbf{Single-WSD}, the model is given a Reddit comment and one target term and decides whether the term is used in its slang sense. This is the simplest setting because the target term is already provided. In \textbf{List-WSD}, the model is given a comment and a list of candidate terms and performs WSD for each term.

\section{Methodology} \label{sec:methodology}

To examine how social-network structure and pragmatic language use shape the diffusion of slang across online communities, this study models the rate at which users adopt new slang as a function of both network and linguistic features. This section describes the features used in the analysis, how they are calculated, and how they capture the social and linguistic mechanisms underlying observed adoption behaviour.

\subsection{Social Network Features}

Prior work suggests that network centrality captures different forms of social capital relevant to language diffusion. \textbf{Degree centrality} has been used as a proxy for \textit{bonding capital}, since it reflects how embedded a user is in their local interaction network \cite{Shin2021-yc}. This aligns with sociolinguistic evidence from the Belfast Study, which found that dense social ties play a central role in dialect maintenance \cite{doi:https://doi.org/10.1002/9781118335598.ch19}. The degree centrality $C_D$ for user $i$ is
$$C_D(i) = \frac{k_i}{n-1}$$
where $k_i$ is the degree of node $i$ and $n$ is the total number of nodes in the network. 

In contrast, weak ties are typically associated with linguistic change rather than maintenance. Sociolinguistic work has argued that users who maintain ties to everyday acquaintances are more likely to facilitate the spread of linguistic innovations across social groups \cite{doi:https://doi.org/10.1002/9781118335598.ch19}. In network terms, these users possess \textit{bridging capital}: they connect otherwise separated parts of the social structure and allow linguistic forms to move beyond tightly bonded local clusters. To capture this mechanism, we include \textbf{Betweenness Centrality} as a feature representing the extent to which slang users occupy bridging positions \cite{Shin2021-yc}. Betweenness centrality measures how often a node sits on the shortest path between other pairs of nodes:
$$C_B(i) = \sum_{s \ne i \ne t} {\frac{\sigma_{st}(i)}{\sigma_{st}}}$$
where $\sigma_{st}$ is the total number of shortest paths from $s$ to $t$, and $\sigma_{st}(i)$ is the number of those paths that pass through $i$. Because raw betweenness centrality depends on network size—that is, as the number of nodes increases, the number of node pairs whose shortest paths could pass through a focal node also increases—we normalize each node’s raw betweenness by the maximum possible betweenness in a network of the same size.
Specifically, normalized betweenness centrality is calculated as $$C'_{B}(i) = \frac{C_{B}(i)}{N_{max}}$$ where $$N_{max} = \frac{(N-1)(N-2)}{2}$$

\subsection{Linguistic Features} \label{linguistic-feature}

Prior work has shown that the linguistic context in which a word is used can influence its likelihood of successful adoption \cite{stewart-eisenstein-2018-making, Altmann2011-hc}. Retrieving slang terms is challenging because slang WSD often involves distinguishing a slang sense from an existing sense that shares the same word form. Recent advances in LLMs allow us to perform this WSD task with high accuracy; see Section \ref{sec:LLM-Detection} for our model evaluation. We leveraged LLMs to identify which glossary terms appear in each utterance and determine whether they are used with the slang senses defined in the glossary. In total, we identified $3,965$ slang terms across $54$ subreddits.

After slang detection, we calculate the linguistic features for each slang term $W_i$. We use Sentence-BERT \cite{reimers-2019-sentence-bert} to embed the $N$ utterances ${U_{W_{i,1}}, U_{W_{i,2}}, \ldots, U_{W_{i,n}}}$ containing $W_i$, obtaining the corresponding embeddings ${E_{W_{i,1}}, E_{W_{i,2}}, \ldots, E_{W_{i,n}}}$. Based on these embeddings, we calculate two semantic features: \textbf{Semantic Dispersion} and \textbf{Effective Topics}.

\paragraph{Semantic Dispersion.} \label{sematnci-dispersion}
We define the Semantic Dispersion as how flexibly a slang term has been adopted across contexts. The $SD$ is calculated for each slang $W_i$ by the mean pairwise cosine distance, formally: $$SD(W_i) = \frac{2}{N(N-1)}\sum_{i < j}{1-\text{cos}(E_{i_j}, E_{i_k})}$$ where $N$ is the number of embeddings and $\text{cos}(\cdot)$ is the cosine similarity. 

\paragraph{Effective Topics.} \label{effective-topic}Drawing the inspiration from \citet{giulianelli-etal-2020-analysing}, we define our Effective Topics to measure how many clusters of topics that a word appeared in. We take embeddings $\{E_{W_{i},1},E_{W_{i},2}...,E_{W_{i},m}\}$ of word $W_i$ and run the K-means algorithm to cluster the sense embeddings. The optimal $K$ is determined by computing the silhouette score to judge how good the embedding fits its assigned cluster. We then calculate the Shannon Entropy $H$ for a word $W_i$ to measure how uncertain on which cluster any of the embeddings will belong to. Mathematically, $$H(W_i) = -\sum_{k=1}^{K}{P_k\text{log}_2 P_k}$$ where $P_k = \frac{n_k}{N}$ is the empirical proportion of usages of word $W_i$ assigned to cluster $k$, with $n_k$ denotes the number of embeddings falling into cluster $k$. To obtain a more interpretable quantity, we exponentiate the entropy to recover the effective number of topics: $$\text{ET}(W_i) = 2^{H(W_i)}$$

\section{Experiments}

\begin{figure*}[t]
  \centering
  \includegraphics[width=1\textwidth]{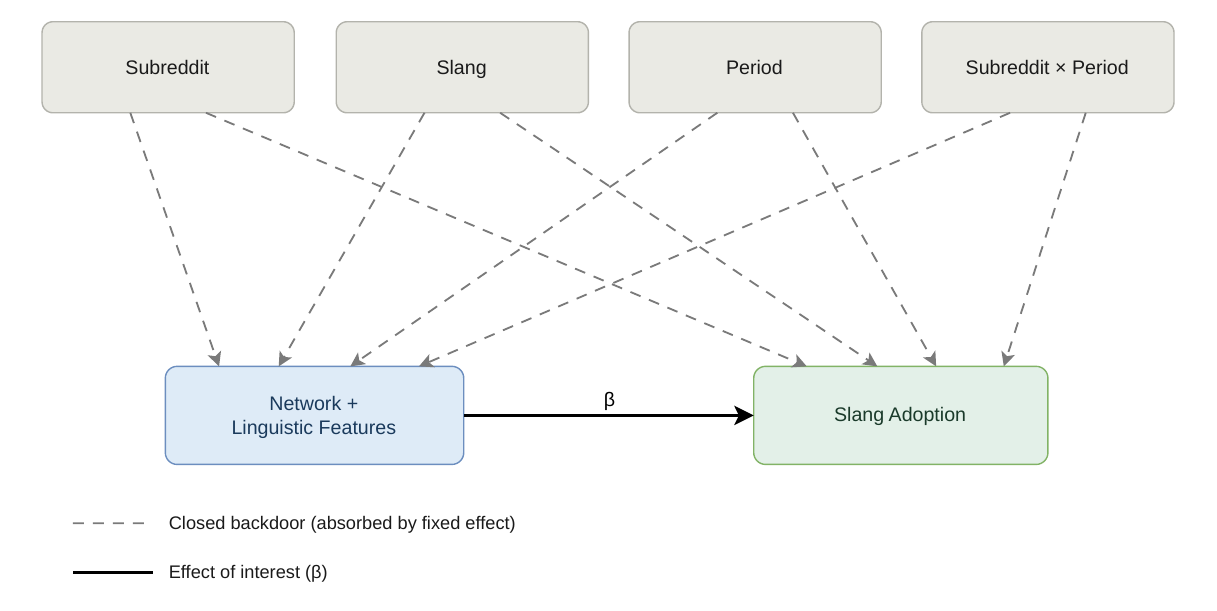}
  \caption{Diagram for fixed effects identification. Dashed gray arrows mark backdoors closed by subreddit, slang, and time fixed effects.}
  \label{fig:dag-threeway-fe}
\end{figure*}

\subsection{LLM Detection} \label{sec:LLM-Detection}

The first step of our analysis is to evaluate LLMs performance on slang detection. Slang detection is challenging for LLMs because they must infer the intended meaning of a term from its surrounding context, particularly when slang reuses existing word forms with different meanings, while also processing potentially long conversational contexts. To better understand the bottlenecks in LLM-based slang detection, we therefore decompose the task into two diagnostic settings:

\paragraph{Single-WSD.} Given a Reddit utterance, one candidate slang term, and its glossary definition, the model decides whether the term is used with the glossary-defined slang sense. This setting isolates word sense disambiguation because the target term is already specified.

\paragraph{List-WSD.} Given a Reddit utterance and the full slang glossary for the subreddit, the model determines which glossary terms are used with their slang meanings. This setting evaluates performance when the model must search over multiple candidate terms while also disambiguating their senses. To ensure a fair WSD setting, for both tasks, we also provide the model with the source subreddit and the conversational context of the utterance, defined as its parent and grandparent comments where available.

\paragraph{Batch Inference.}
During LLM-based slang detection, we used vLLM \cite{kwon2023efficient} for batched inference on a single GH200 GPU, with the temperature set to 0 and top\_p set to 1. To reduce computational cost, we first used string matching to identify utterances containing candidate glossary terms. For each candidate utterance, we provided the model with only the corresponding subset of glossary entries, together with the surrounding context, and asked it to determine whether each candidate term was used with the slang sense defined in the glossary. This pre-filtering step was designed solely to improve inference efficiency. With this setup, processing 10,000 samples took approximately 4 minutes.

After LLM inference, we performed a separate post-hoc cleaning step to improve detection precision. Because our analysis focuses only on terms included in the glossary, and LLMs may hallucinate terms that do not appear in the original Reddit comment \cite{yang2026hallucinatelongresponsegeneration}, we used string matching to remove detected terms that were either absent from the glossary or not present in the original utterance. This post-processing step therefore reduces false positives in the final slang detections.

\paragraph{Results.}
We evaluate four open-source language models: Llama-3.1-8B-Instruct \cite{grattafiori2024llama3herdmodels}, Olmo-3-7B-Instruct \cite{olmo2026olmo3}, Ministral-3-8B-Instruct \cite{liu2026ministral3}, and Gemma-4-E4B-it \cite{gemmateam2026gemma4technicalreport}, on our human-annotated Reddit utterances. We restrict our evaluation to small, open-source models because our downstream analysis targets roughly 100 million utterances, at which scale inference cost dominates model choice. 

As shown in Table~\ref{tab:benchmark-result}, Gemma-4-E4B-it performs best on both tasks, achieving $93.44\%$ F1 on Single-WSD and $83.98\%$ F1 on List-WSD. The main bottleneck therefore appears not to be slang WSD itself, but rather processing long contexts while reliably following output instructions. List-WSD is particularly challenging because the prompt includes the glossary and surrounding context. Under this longer context, models often generate extra commentary, which contaminates outputs and increases inference time.

Based on these results, we adopt Gemma-4-E4B-it for slang detection in the remainder of the analysis.

\subsection{Slang Level Analysis} \label{exp-1}

Having established that LLM-based slang detection is sufficiently reliable, we use the LLM-detected data to compute both network and linguistic features and investigate how they predict the adoption rate. Following prior work on temporal network analysis \cite{zhu-jurgens-2021-structure, Hamilton2017-cy}, we partition each subreddit's continuous interaction stream into monthly windows. For each subreddit $S_i$ with data spanning $T_i$ months, we obtain a sequence of monthly interaction networks $\{I_{S_i, 1}, I_{S_i, 2}, \dots, I_{S_i, T_i}\}$. We operationalize an interaction between two users as occurring when the two users are in close proximity, separated by at most two comments \cite{Hamilton2017-cy, zhu-jurgens-2021-structure}.

For each window $I_{S_i, t}$, we compute both network and linguistic features and use them to predict the number of new adopters in the \emph{following} window $I_{S_i, t+1}$. This lagged design lets us probe how well current-window features explain future adoption, rather than merely correlating with concurrent usage. Rather than aggregating across all slang terms within a subreddit, we model adoption at the slang-term level to preserve term-level variation in the data. For each window $I_{S_i, t}$, let $\{s_1, s_2, \dots, s_{n_{i,t}}\}$ denote the set of slang terms observed in that window. For each slang term $s_k$, the response variable $y_{i,k,t}$ is defined as the number of new adopters of $s_k$ in $I_{S_i, t+1}$.

\paragraph{Network features.}

For each slang term $s_k$ in window $I_{S_i, t}$, let $\mathcal{U}_{i,k,t}$ denote the set of users who used $s_k$ during that window, with $|\mathcal{U}_{i,k,t}| = M_{i,k,t}$. For each user $u \in \mathcal{U}_{i,k,t}$, we compute two centrality measures on $I_{S_i, t}$: degree centrality, and betweenness centrality in the complete subreddit interaction network $I_{S_{i},t}$. We then average each measure across the $M_{i,k,t}$ adopters to obtain a slang-level network feature vector $\mathbf{x}^{\text{net}}_{i,k,t}$:
$$\mathbf{x}^{\text{net}}_{i,k,t} = \frac{1}{M_{i,k,t}} \sum_{u \in \mathcal{U}_{i,k,t}} \mathbf{c}(u; I_{S_i, t}),$$
where $\mathbf{c}(u; I_{S_i, t})$ stacks the two centrality scores for user $u$ in network $I_{S_i, t}$.

\paragraph{Linguistic features.}
To compute linguistic features for $s_k$ at window $t$, we sample 100 utterances containing $s_k$ from $I_{S_i, t}$ and encode each utterance with SentenceBERT. From the resulting 100 embeddings, we compute Semantic Dispersion and Effective Topics as described in Section~\ref{linguistic-feature}, yielding a two-dimensional linguistic feature vector $\mathbf{x}^{\text{ling}}_{i,k,t}$.

\paragraph{Combined covariates.}
Concatenating network and linguistic features gives the full covariate vector for each (subreddit, slang, window) triple:
$$\mathbf{X}_{i,k,t} = \left[\, \mathbf{x}^{\text{net}}_{i,k,t} \;\Vert\; \mathbf{x}^{\text{ling}}_{i,k,t} \,\right],$$
which is used to predict $y_{i,k,t}$.

\paragraph{Model.}
Prior work has examined how network features and linguistic features each predict slang diffusion in isolation \cite{zhu-jurgens-2021-structure}. They explicitly note that external surges in adoption are essentially undetectable from the interaction data alone, yet may substantially shape adoption dynamics.

For example, in \texttt{r/asoiaf} (the subreddit devoted to George R. R. Martin's \emph{A Song of Ice and Fire}), new slang tends to emerge whenever a new book is released: the term \emph{ASOS}, for instance, was coined and spread rapidly following the publication of \emph{A Storm of Swords}. A different pattern played out in r/wallstreetbets during the 2021 GameStop short squeeze, when a massive influx of new users coincided with a sharp acceleration in the subreddit's slang adoption rate. In both cases, an external event simultaneously shifted the network structure, the linguistic environment, and the rate of slang adoption, making it impossible to attribute adoption to any single feature without accounting for the underlying event.

To minimize the external effects, Figure~\ref{fig:dag-threeway-fe} summarizes the sources of confounding we account for. We posit that three classes of factors jointly drive both the features and adoption:
\begin{itemize}
    \item \textbf{Subreddit factors:} community culture, topic focus, moderation style, and stable demographic composition. Figure \ref{fig:r/boxoffice} illustrates how an exogenous event can affect slang usage.
    \item \textbf{Slang factors:} properties intrinsic to the slang term itself, including length, phonology, semantic class, and age. Figure \ref{fig:r/cars} shows that slang terms with more common, everyday meanings may be used more frequently than other terms, such as \texttt{SUV} compared with \texttt{cam}.
    \item \textbf{Period factors:} Reddit-wide trends, platform changes, and viral slang moments that affect all communities simultaneously. Figure \ref{fig:r/clashroyale} shows that when a new game mode was released, usage of the corresponding slang term increased sharply and then stabilized. Another source of variation is the overall growth in subreddit activity showed in Figure \ref{fig:r/utteranceall}, which can indirectly increase the observed frequency of slang usage.
    \item \textbf{Subreddit $\times$ Period interactions:} Subreddit-specific events at particular moments, such as community drama, moderation changes, or sudden influxes of users. Figure \ref{fig:r/AFL} shows that \texttt{r/AFL} exhibits seasonal patterns in both overall utterance volume and slang usage across the year.
\end{itemize}

We estimate a fixed-effects negative binomial count model with a log link, where the outcome is the number of new adopters of a slang term in the following month. We include the log number of users at risk of adoption as an offset and compute cluster-robust standard errors at the subreddit level.

\begin{table*}[t]
  \centering
  \small
  \setlength{\tabcolsep}{7pt}
  \renewcommand{\arraystretch}{1.35}
  \begin{tabular}{lrrrr}
    \textbf{Feature} 
      & \textbf{Estimate} 
      & \textbf{Std. Error} 
      & \textbf{$z$ value} 
      & \textbf{$p$ value} \\
    \hline

    Mean Betweenness
      & $0.103^{*}$
      & $0.052$
      & $1.989$
      & $0.047$ \\

    Mean Degree
      & $-0.506^{***}$
      & $0.137$
      & $-3.704$
      & $<0.001$ \\

    Effective Topics
      & $-0.018^{***}$
      & $0.003$
      & $-5.804$
      & $<0.001$ \\

    Semantic Dispersion
      & $-0.215^{***}$
      & $0.053$
      & $-4.089$
      & $<0.001$ \\

  \end{tabular}
  \caption{Regression Coefficients for Predicting Slang Adoption. 
  Statistical significance is denoted by $^{*}p<0.05$, 
  $^{**}p<0.01$, and $^{***}p<0.001$.}
  \label{tab:regression-coefficients}
\end{table*}

\paragraph{Controlling for confounding with fixed effects.}
To account for these shared drivers, we include fixed effects for subreddit, time period, subreddit-period interaction, and slang term. These fixed effects control for systematic differences across communities, periods, and individual slang terms. For example, when examining slang used in \texttt{r/wow} in March 2026, the subreddit and time are held fixed, while slang-term fixed effects account for each term’s baseline tendency to be adopted. The model then asks whether differences in Effective Topics are associated with differences in subsequent adoption, rather than attributing those differences to the community, time period, or slang term itself. In causal-inference terms, these fixed effects help block back-door paths from shared contextual factors to slang adoption \citep{Huntington-Klein2021-jt}.

\begin{figure*}[t]
  \centering
  \includegraphics[width=1\textwidth]{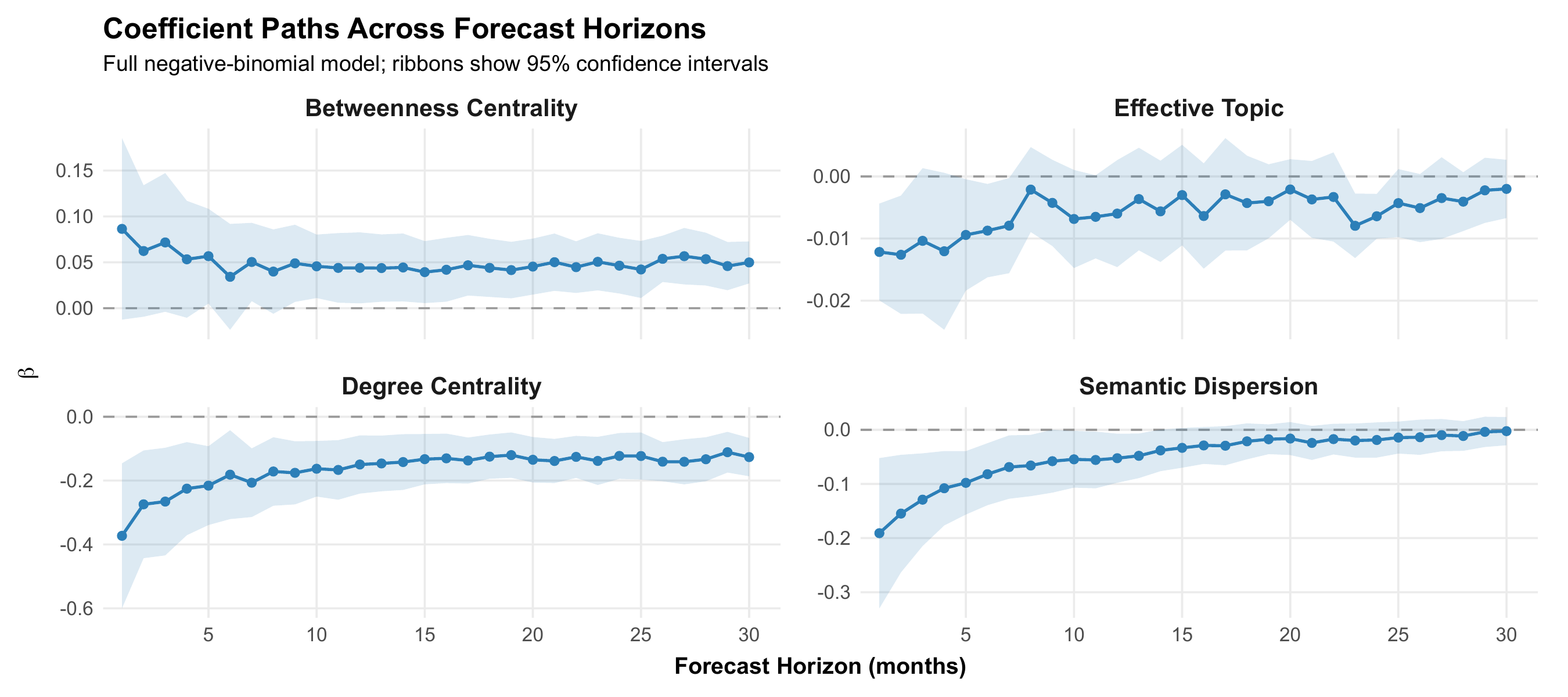}
  \caption{Shows how the estimated effects of the four covariates change across forecast horizons. Betweenness centrality remains consistently positive, while degree centrality remains negative across all horizons. In contrast, the negative effects of effective topics and semantic dispersion gradually attenuate toward zero as the forecast horizon increases.}
  \label{fig:coefficient}
\end{figure*}

\paragraph{Results.} 
Table \ref{tab:regression-coefficients} shows the results of fitting our fixed-effects model. Looking at the covariate estimates, betweenness centrality shows a positive effect on new adoption $(\beta = 0.103, SE = 0.052, z = 1.98, p < 0.05)$, while degree centrality shows the largest negative effect $(\beta = -0.506, SE = 0.137, z = -3.704, p < 0.001)$. Both effective topics $(\beta = -0.018, SE = 0.003, z = -5.804, p < 0.001)$ and semantic dispersion $(\beta = -0.215, SE = 0.053, z = -4.089, p = 0.001)$ show negative effects.

We also assessed robustness to LLM classification error by using sensitivity and specificity, adjusted for observed prevalence, to Monte Carlo-simulate corrected adoption outcomes and refit the model. Table \ref{tab:monte-carlo-robustness} showed all coefficients retained their signs and significance across simulations.

The network features therefore appear to play an important role in slang adoption than the linguistic features. Following the interpretation introduced earlier, we treat betweenness centrality as a proxy for bridging positions across otherwise weakly connected parts of the network, and degree centrality as a proxy for local connectedness and embeddedness. The positive coefficient on betweenness is consistent with Milroy and Milroy's account of the role of weak ties in linguistic diffusion \cite{doi:https://doi.org/10.1002/9781118335598.ch19}: when adopters have higher average betweenness, they are more likely to occupy bridging positions that connect different parts of the network, creating pathways through which slang can spread beyond local clusters. Degree centrality shows the opposite pattern. When adopters have higher average degree and are more densely connected within their local network, subsequent slang adoption decreases. Together, these results suggest that slang diffusion may benefit more from connections that bridge otherwise separated parts of a community than from dense local connectivity alone.

The linguistic features tell a more complex story. Slang used across more diverse contexts or topics is associated with fewer new adopters, possibly because such variation makes the slang harder for potential adopters to understand and use confidently. To examine whether this effect persists over time, we fit the same model to predict new adopters at \(t+n\). As shown in Figure \ref{fig:coefficient}, the negative effect of semantic dispersion gradually approaches zero across longer horizons, while the centrality effects remain relatively stable. A Wald test confirms that only semantic dispersion changes significantly from \(t+1\) to \(t+30\). This suggests that contextual diversity may create a temporary barrier to slang adoption that gradually fades over time.


\subsection{User Level Analysis}

\begin{table*}[t]
  \centering
  \small
  \setlength{\tabcolsep}{10pt}
  \renewcommand{\arraystretch}{1.35}
  \begin{tabular}{lcccc}
    \textbf{Variable} 
      & \textbf{$\beta$} 
      & \textbf{HR} 
      & \textbf{95\% CI} 
      & \textbf{$p$-value} \\
    \hline

    Degree Centrality
      & $0.190$
      & $1.209$
      & $[1.131, 1.292]$
      & $<.001$ \\

    Betweenness Centrality
      & $-0.123$
      & $0.884$
      & $[0.793, 0.986]$
      & $.027$ \\

    Betweenness Centrality $\times \log(t)$
      & $-0.055$
      & $0.946$
      & $[0.880, 1.018]$
      & $.137$ \\

    Degree Centrality $\times \log(t)$
      & $-0.012$
      & $0.989$
      & $[0.951, 1.027]$
      & $.559$ \\

    Semantic Dispersion
      & $0.057$
      & $1.059$
      & $[1.020, 1.099]$
      & $.003$ \\

    Effective Topics
      & $-0.077$
      & $0.926$
      & $[0.890, 0.963]$
      & $<.001$ \\

  \end{tabular}
  \caption{Cox proportional hazards model results. Time-varying effects are modeled as interactions with $\log(t)$.}
  \label{tab:cox-table}

\end{table*} \label{table-3}
In the Slang Level Analysis, we examined whether the social structure of slang adopters and slang semantics can predict the number of new adopters in a future time window. We found that network structure and slang semantics are strongly associated with future adoption. From the other side of the diffusion process, one question is whether the same set of features predicts when users will adopt a slang term? In this part of analysis, we used Cox Proportional Hazards (CPH) model \cite{https://doi.org/10.1002/bimj.4710310209} to incorporate both network and linguistic features to predict the time from a user's first encounter with a slang term to their first use of that term.

\paragraph{Data.}
In this experiment, we shift to user-level analysis of slang adoption. For each subreddit $k$, slang term $s_k$, and time window $I_{S_k, t}$, let $\{u_1, u_2, \ldots, u_n\}$ denote the users who posted at least one utterance in $I_{S_k, t}$. For each user, we record two timestamps: $t_{\text{encounter}}$, the time at which the user first encountered $s_k$, and $t_{\text{adopt}}$, the time at which the same user first used $s_k$ themselves.

We use the same definition of interaction as in Section~\ref{exp-1}. $u_i$ is said to encounter $s_k$ at time $t$ if $u_i$ posts a comment at time $t$ that is separated by at most two comments to a comment $c$ that contains $s_k$. A user is considered an adopter of $s_k$ if they subsequently post a comment containing $s_k$ denoted as $t_{\text{adopt}}$. Users who encounter $s_k$ but never post it themselves are treated as right-censored in the survival analysis. The resulting dataset contains tuples of the form $\{u_i, k, s_k, t_{\text{encounter}}, t_{\text{adopt}}\}$, with $t_{\text{adopt}} = \infty$ for censored users.

For predictive features, we reuse the set studied in Section~\ref{sec:methodology}: degree centrality and betweenness centrality as network features, and semantic dispersion and effective topics as linguistic features. All features are computed for user $u_i$ at each time point from $t_{\text{encounter}}$ until $t_{\text{adopt}}$, or through the end of the observation period if the user is right-censored.

\paragraph{Model.}
We use a Cox model allowing non-proportional hazards \cite{Harrell2001-vh} to estimate how network and linguistic features affect the timing of slang adoption. Given a set of users at risk of adoption, the model estimates the hazard function:
$$h(t \mid X) = h_{0,\text{group}}(t)\, \exp(\beta(t)^\top X),$$
where $h(t \mid X)$ is the instantaneous adoption hazard for a user with covariate vector $X$ at time $t$ and $\beta$ is the vector of potentially time-varying log-hazard coefficients for the covariates. We report exponentiated coefficients as hazard ratios: a ratio above $1$ indicates that the covariate accelerates adoption, while a ratio below $1$ indicates that it suppresses adoption. The model assumes that the effect of each covariate is constant over time but Schoenfeld-residual diagnostics indicated potential non-proportionality for degree and betweenness centrality; we therefore included interactions with \(\log(t)\) for these covariates.

\paragraph{Results.}

The Cox proportional hazards results show a nuanced picture of individual slang adoption. Degree centrality is associated with faster slang adoption, whereas betweenness centrality is associated with slower adoption. Users with higher degree Centrality are more strongly embedded in the community and may therefore have more opportunities to encounter and learn its conventional language, making them faster to adopt a slang term once it begins circulating.

Betweenness centrality provides an important distinction between the ability to transmit information and the tendency to adopt it. Users with high betweenness occupy brokerage positions connecting otherwise separated parts of the network, which may make them particularly important for spreading a slang term once they have adopted it. However, occupying such a bridge position does not inherently imply a greater propensity to accept new linguistic conventions. Indeed, our Cox model shows the opposite association: users with higher betweenness adopt more slowly. One possible explanation is that brokers participate across more heterogeneous conversational environments and are consequently less embedded in the linguistic norms of any single local community. Thus, the network positions that are advantageous for carrying a term across communities need not be the same positions that facilitate rapid adoption. Exposure, adoption, and subsequent transmission may therefore represent related but distinct stages of the diffusion process.

The linguistic features reveal a similarly nuanced pattern. Semantic Dispersion is associated with faster individual adoption, whereas Effective Topics is associated with slower adoption. Although both features capture aspects of contextual breadth, they may characterize different forms of contextual variation. Greater semantic dispersion may provide users with varied evidence about how a slang term can be used, supporting the development of a more flexible representation of its meaning. In contrast, a term distributed across many distinct topical contexts may provide more fragmented or less coherent evidence, making it harder for a new user to infer a stable meaning and confidently adopt the term.

This distinction is broadly consistent with mixed findings in the word-learning literature. Contextual diversity has been associated with more efficient lexical processing and earlier vocabulary acquisition \cite{Adelman2006-jg, Jimenez2022-zv}, while recent experimental work has shown that encountering a novel word across multiple disconnected narrative contexts can disrupt the earliest stages of meaning learning \cite{Hulme2023-kp}. \citet{Mak2021-jb} reconcile these findings through an anchoring account: encountering a word initially in a restricted and coherent context can establish a stable representation, after which experience across more diverse contexts may enrich that representation. Our findings suggest a related distinction in slang adoption: contextual versatility may facilitate learning when different usages can be integrated into a coherent representation, whereas fragmentation across many topical environments may impede adoption.

\section{Conclusion}

In this study, we introduced a new benchmark for evaluating slang detection in real-world online social media text. We then used LLM-based slang detection to enable more fine-grained analyses of slang diffusion than was previously feasible. Building on this capability, we presented two analyses of slang use on Reddit. By combining social network and linguistic features, we found that slang diffusion is supported by adopters who occupy bridging positions in the network, whereas greater bonding capital among existing adopters is negatively associated with subsequent diffusion. At the individual level, however, users with greater degree centrality adopt slang more quickly, while users occupying high-betweenness brokerage positions adopt more slowly, suggesting that network positions that facilitate transmission are not necessarily the same as those that facilitate adoption.

Our results also show that contextual diversity is not a unitary property of slang diffusion. Greater semantic dispersion is associated with faster individual adoption, whereas usage across a larger number of distinct topics is associated with slower adoption. Together, these findings suggest that varied semantic usage may provide useful evidence for learning how a slang term can be used, while fragmentation across distinct topical contexts may make its meaning more difficult to establish. More broadly, our findings highlight that both the social structure through which slang travels and the linguistic contexts in which it appears shape how slang spreads through online communities.

\section*{Limitations}

Although we use fixed-effects models to block several backdoor paths and further analyze adoption with Cox models, we are still unable to make causal interpretations from this observational data. Important confounders may remain uncontrolled, such as users’ interests, activity levels, and prior familiarity with the slang terms.

Our interpretation of network features as forms of bonding and bridging social capital is also necessarily approximate. Prior work has offered different interpretations of bonding and bridging capital \cite{Shin2021-yc}, and these concepts do not map one-to-one onto any single network statistic. For example, dense local connectivity, degree centrality, PageRank, and betweenness centrality may each capture different aspects of social embeddedness or brokerage.

We used LLM-detected slang usages in our analysis, so classification errors introduced by the LLM may propagate to our downstream results. We therefore conducted a Monte Carlo sensitivity analysis (Table \ref{tab:monte-carlo-robustness}) to evaluate the robustness of our findings to plausible detection errors. We used the observed prevalence together with the estimated sensitivity and precision to derive prevalence-adjusted misclassification rates for the simulation. However, due to limited manually annotated data, we assume that the estimated detection performance of the LLM is transferable across subreddits. This assumption may not hold if slang usage or detection difficulty systematically differs across communities. As LLM-based slang detection continues to improve, we expect this source of measurement error to become less consequential in future work.

Following prior work such as \citet{zhu-jurgens-2021-structure}, our definition of adoption captures observable production rather than learning or comprehension. We count a user as an adopter only when they use the term in the relevant sense, which allows us to avoid treating mere exposure as adoption. However, users may learn or understand a term without producing it themselves. This distinction is important because contextual diversity may affect comprehension and production differently. For example, prior work suggests that encountering a word in more diverse contexts may not always facilitate early learning and may even be detrimental during the initial anchoring phase \cite{Li2024-fp}. Our results therefore speak specifically to the production of subreddit-specific lexical items, rather than to lexical learning more broadly.

Finally, our analysis is also limited to slang terms included in the glossary, which predominantly contains slang that became sufficiently established to be documented. As a result, our findings primarily characterize the diffusion of relatively successful slang terms rather than slang innovations in general. Slang terms that emerged but failed to spread widely are likely underrepresented or absent from our data, so our conclusions should not be generalized to all newly coined slang.

\section*{Ethics Statement}
\paragraph{Risks.} This research is based on data-driven analysis of publicly available online data. The study does not involve human participants and is not intended to introduce direct harm.

\paragraph{Offensive Content.} We analyzed Reddit data that may contain explicit or offensive language. Because the released dataset contains naturally occurring Reddit conversations, it may also include such content. Researchers using the dataset should therefore be aware that it may contain both explicit language and potentially identifiable information originating from public Reddit posts.

\paragraph{Licenses.}
We use ConvoKit \cite{Chang2020-gr} for Reddit data under the MIT License. We also use vLLM \cite{kwon2023efficient}, an open-source library for LLM inference released under the Apache License 2.0. We use both ConvoKit and vLLM only for research purposes and in accordance with their respective licenses.

\paragraph{AI Assistance.}
We used an AI assistant for grammar checking and language polishing.

\section*{Acknowledgments}
This work used Delta and DeltaAI at the National Center for Supercomputing Applications (NCSA) through allocation HUM240002, CIS260828 from the Advanced Cyberinfrastructure Coordination Ecosystem: Services \& Support (ACCESS) program \cite{access}, which is supported by U.S. National Science Foundation grants \#2138259, \#2138286, \#2138307, \#2137603, and \#2138296. We also used computational resources provided by the Beehive cluster at Toyota Technological Institute at Chicago.


\bibliography{custom}

\clearpage

\clearpage

\appendix

\section{LLM Prompts}

\subsection{List Slang Sense Disambiguation}

We use the following prompt for list slang sense
disambiguation:

\begin{lstlisting}[style=prompt]
{
  "role": "system",
  "content": "You are a precise slang sense disambiguation assistant. Your job is to determine which terms from a provided {{SUBREDDIT}} glossary are used in their specific slang sense within a user's utterance."
},
{
  "role": "user",
  "content": "GLOSSARY (authoritative; acts as your candidate list):\n{{GLOSSARY}}\n\nCONTEXT (for disambiguation only):\n{{CONTEXT}}\n\nUTTERANCE:\n{{UTTERANCE}}\n\nTASK:\nIdentify all terms from the provided GLOSSARY that appear in the UTTERANCE and are used according to their specific {{SUBREDDIT}} slang definition.\n\nDECISION RULES:\n1) Evaluate every term defined in the GLOSSARY against the UTTERANCE.\n2) Use the UTTERANCE and CONTEXT to determine the intended meaning of each term.\n3) If a glossary term is used in the UTTERANCE with the exact slang/jargon meaning defined in the GLOSSARY, include it in your output.\n4) If a glossary term is used with a literal, standard, or different meaning, exclude it.\n\nOUTPUT FORMAT (STRICT):\nReturn a comma-separated list of the exact terms from the glossary that are used in their slang sense (e.g., term1, term2).\nIf no terms from the glossary are used in their slang sense, return exactly:\n[NO SLANG]\n\nDo NOT include explanations, punctuation (other than commas separating terms), JSON, or any extra text."
}
\end{lstlisting}

\subsection{Single Slang Sense Disambiguation}

We use the following prompt for single slang sense
disambiguation:

\begin{lstlisting}[style=prompt]
{
  "role": "system",
  "content": "You are a precise slang sense disambiguation assistant. Your job is to determine whether a given candidate term appearing in a user's utterance is used with the {{SUBREDDIT}} slang meaning defined in a provided glossary."
},
{
  "role": "user",
  "content": "GLOSSARY (authoritative):\n{{GLOSSARY}}\n\nCONTEXT (for disambiguation only):\n{{CONTEXT}}\n\nUTTERANCE:\n{{UTTERANCE}}\n\nCANDIDATE TERM (appears in the utterance):\n{{TERM}}\n\nTASK:\nDetermine whether the CANDIDATE TERM in the UTTERANCE is used with the {{SUBREDDIT}} slang meaning defined in the GLOSSARY.\n\nDECISION RULES:\n1) The candidate term must correspond to a glossary entry (base form match allowed).\n2) Use the UTTERANCE and CONTEXT to determine the meaning.\n3) If the candidate term is used with the slang/jargon meaning defined in the glossary, return True.\n4) If the candidate term is used with a literal, standard, or different meaning, return False.\n5) Only evaluate the provided candidate term. Do NOT extract or evaluate other words.\n\nOUTPUT FORMAT (STRICT):\nReturn exactly one token:\nTrue\nor\nFalse\n\nDo NOT include explanations, punctuation, JSON, or extra text."
}
\end{lstlisting}

\subsection{Human-Annotated Benchmark Statistics}

Table~\ref{tab:benchmark_statistics} summarizes the
human-annotated benchmark used for evaluating slang
sense detection.

\begin{table}[h]
    \centering
    \small
    \setlength{\tabcolsep}{5pt}
    \renewcommand{\arraystretch}{1.15}
    \begin{tabular}{lr}
        \textbf{Statistic} & \textbf{Count} \\
        \hline
        Number of utterances
            & 1,333 \\
        Positive utterances
            & 613 \\
        Negative utterances
            & 720 \\
        Glossary-term occurrences
            & 1,133 \\
        Occurrences in glossary senses
            & 961 \\
        Occurrences in non-glossary senses
            & 172 \\
        Unique glossary terms evaluated
            & 345 \\
    \end{tabular}
    \caption{Summary statistics of the human-annotated slang sense disambiguation benchmark. Note that a single utterance may contain multiple slang instances; therefore, the 613 slang-containing utterances include 1,333 slang terms in total. All three authors independently annotated the same 50 samples using the same materials: the glossary and utterance context, with access to the Internet when needed to determine the meaning. Agreement among the three annotators yielded a Fleiss’ $\kappa$ of $0.783$, indicating very high inter-rater agreement.
}
    \label{tab:benchmark_statistics}
\end{table}

\section{Additional Results}
\label{sec:additional_results}

This section provides additional plots and figures from our analyses.

\clearpage

\begin{table*}[t]
  \centering
  \scriptsize
  \setlength{\tabcolsep}{4pt}
  \renewcommand{\arraystretch}{1.25}
  \begin{tabular}{lrrrrrr}
    \textbf{Feature} 
      & \shortstack{\textbf{Orig.} \\ $\boldsymbol{\beta}$}
      & \shortstack{\textbf{Median} \\ $\boldsymbol{\beta}$}
      & \shortstack{\textbf{2.5\%} \\ \textbf{Quant.}}
      & \shortstack{\textbf{97.5\%} \\ \textbf{Quant.}}
      & \shortstack{\textbf{Same} \\ \textbf{Sign}}
      & \shortstack{\textbf{Prop.} \\ $\boldsymbol{p<.05}$} \\
    \hline

    Mean Betweenness
      & $0.103$
      & $0.074$
      & $0.069$
      & $0.078$
      & $1.00$
      & $1.00$ \\

    Mean Degree
      & $-0.506$
      & $-0.347$
      & $-0.353$
      & $-0.339$
      & $1.00$
      & $1.00$ \\

    Effective Topics
      & $-0.018$
      & $-0.004$
      & $-0.005$
      & $-0.003$
      & $1.00$
      & $1.00$ \\

    Semantic Dispersion
      & $-0.215$
      & $-0.064$
      & $-0.065$
      & $-0.063$
      & $1.00$
      & $1.00$ \\

  \end{tabular}
  \caption{Monte Carlo robustness analysis accounting for LLM classification error. Orig. $\beta$ denotes the original coefficient estimate, while Median $\beta$ denotes the median coefficient across simulations. Same Sign reports the proportion of simulations with the same coefficient direction as the original estimate, and Prop. $p<.05$ reports the proportion that remain statistically significant.}
  \label{tab:monte-carlo-robustness}
\end{table*}

\begin{figure*}[t]
    \centering
    \includegraphics[width=\linewidth]
    {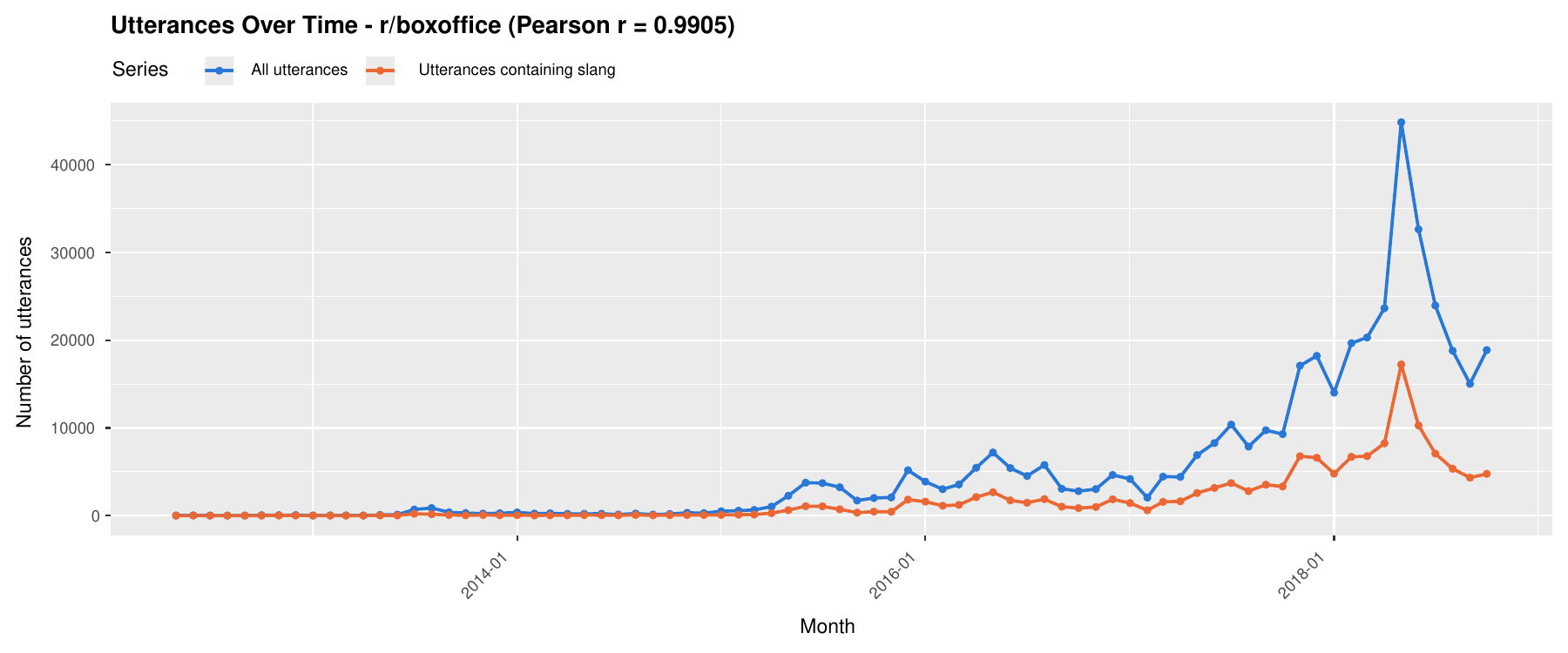}
    \caption{This figure shows the total number of utterances in \texttt{r/boxoffice} over time. The sharp increase in activity in April 2018 coincides with the release of \texttt{Avengers: Infinity War}. This exogenous event substantially increased subreddit activity, potentially affecting both the structure of the interaction network and the frequency of observed slang usage. Pearson's $r$ shows a high correlation between the number of slang utterances and the total number of utterances.}
    \label{fig:r/boxoffice}
\end{figure*}

\begin{figure*}[t]
    \centering
    \includegraphics[width=\linewidth]
    {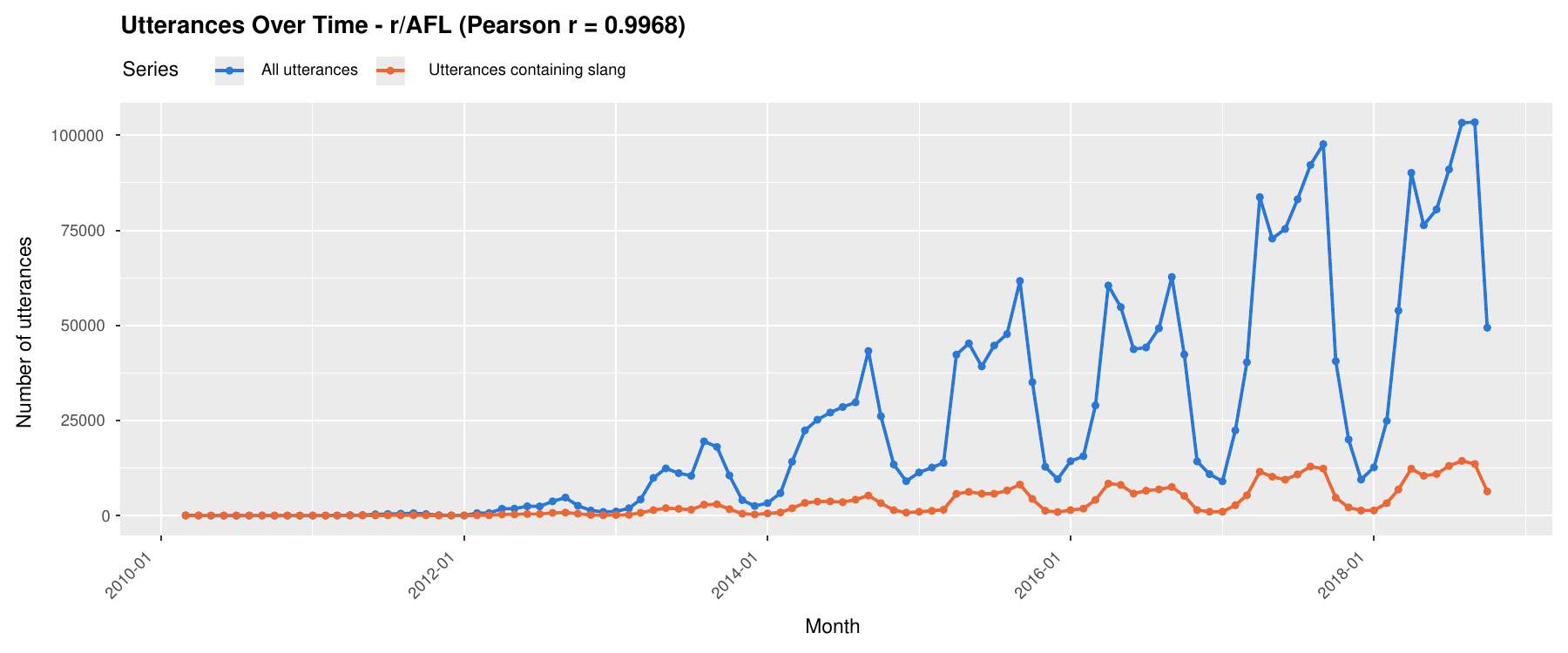}
    \caption{This figure shows the total number of utterances in \texttt{r/AFL} over time. There are systematic fluctuations throughout the year due to the seasonal structure of the AFL competition.}
    \label{fig:r/AFL}
\end{figure*}

\begin{figure*}[t]
    \centering
    \includegraphics[width=\linewidth]
    {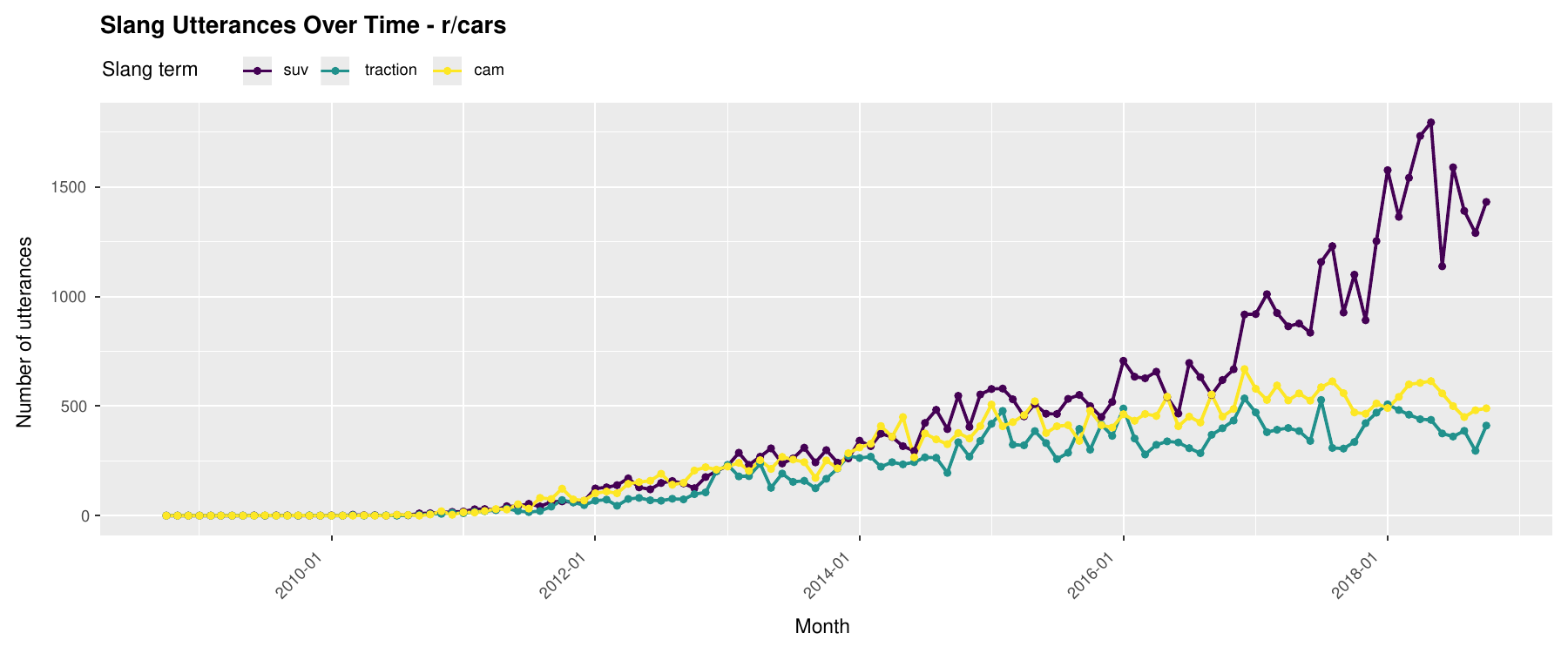}
    \caption{This figure shows the usage trajectories of the slang terms \texttt{SUV}, \texttt{traction}, and \texttt{cam} in \texttt{r/cars}. The \texttt{SUV} refers to a more popular type of car and is used more frequently in the community, resulting in a substantially different growth trajectory from those of the other slang terms.}
    \label{fig:r/cars}
\end{figure*}

\begin{figure*}[t]
    \centering
    \includegraphics[width=\linewidth]
    {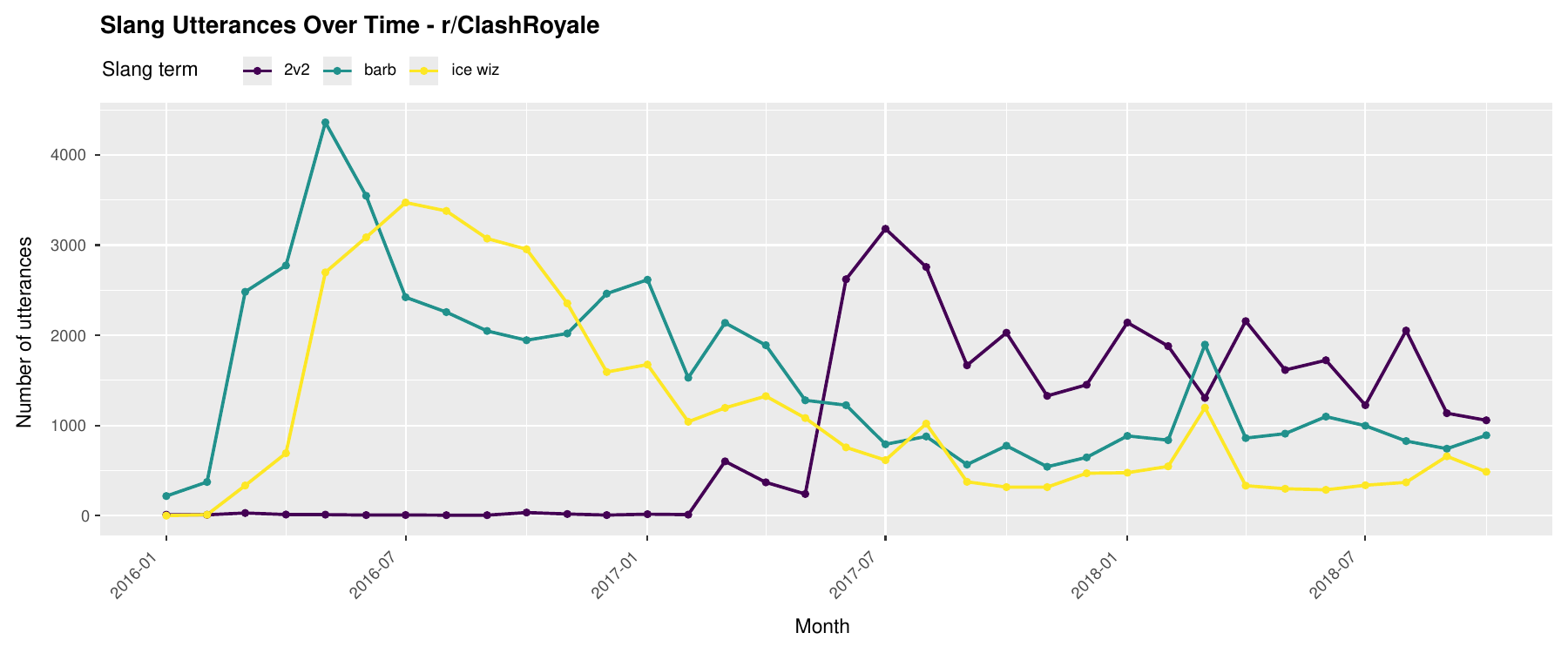}
    \caption{This figure shows the usage trajectories of the slang terms \texttt{2v2}, \texttt{barb}, and \texttt{ice wiz} in \texttt{r/ClashRoyale}. The \texttt{2v2} game mode was released in March 2017, after which the usage of the term remained relatively stable.}
    \label{fig:r/clashroyale}
\end{figure*}

\begin{figure*}[t]
    \centering
    \includegraphics[width=\linewidth]
    {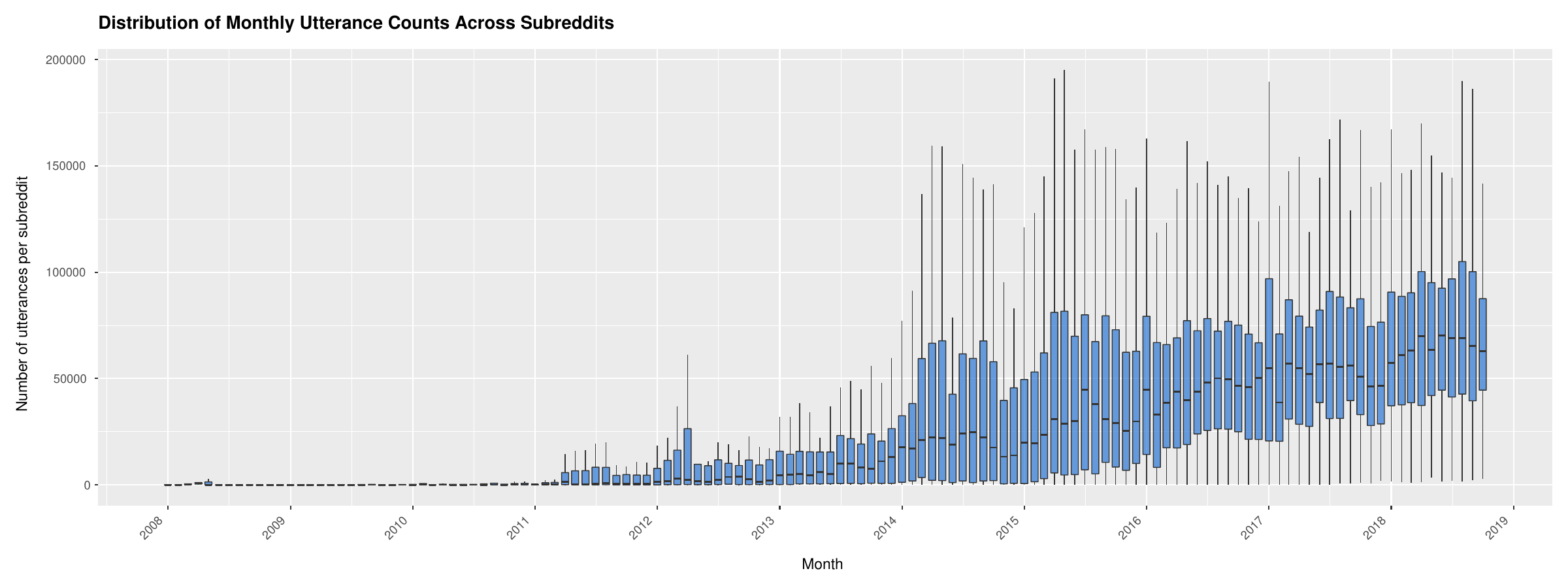}
    \caption{This figure shows the distribution of the number of utterances across all 54 subreddits over time. There is a general upward trend in Reddit activity, likely reflecting the growth of the Reddit community and increased access to the internet.}
    \label{fig:r/utteranceall}
\end{figure*}

\end{document}